\documentclass[10pt,twocolumn,letterpaper]{article}

\usepackage[pagenumbers]{cvpr}

\usepackage{amsmath}
\usepackage{amssymb}
\usepackage{booktabs}
\usepackage{multirow}
\usepackage{algorithm}
\usepackage{algorithmic}
\usepackage{enumitem}
\usepackage{subcaption}
\usepackage{microtype}

\definecolor{cvprblue}{rgb}{0.21,0.49,0.74}
\usepackage[pagebackref,breaklinks,colorlinks,allcolors=cvprblue]{hyperref}

\def\paperID{*****}
\def\confName{CVPR}
\def\confYear{2026}

\title{Sketch2Feedback: A Grammar-in-the-Loop Framework for\\Rubric-Aligned Feedback on Student STEM Diagrams}

\author{Aayam Bansal\\
IEEE\\
{\tt\small aayambansal@ieee.org}}

\begin{document}
\maketitle

\begin{abstract}
Providing timely, rubric-aligned feedback on student-drawn diagrams is a persistent challenge in STEM education.
While large multimodal models (LMMs) can jointly parse images and generate explanations, their tendency to hallucinate undermines trust in classroom deployments.
We present \textit{Sketch2Feedback}, a grammar-in-the-loop framework that decomposes the problem into four stages---hybrid perception, symbolic graph construction, constraint checking, and constrained VLM feedback---so that the language model verbalizes \emph{only} violations verified by an upstream rule engine.
We evaluate on two micro-benchmarks, FBD-10 (free-body diagrams, 200 samples) and Circuit-10 (circuit schematics, 200 samples), comparing our pipeline (Qwen2-VL-2B) against an end-to-end LMM (LLaVA-1.5-7B) and a vision-only detector.
Results are mixed and instructive: the end-to-end LMM achieves stronger error detection on free-body diagrams (micro-F1 0.471 \vs 0.263), while the grammar pipeline substantially outperforms on circuits (0.329 \vs 0.038) and attains perfect actionability (5.0/5).
The grammar pipeline's high circuit hallucination rate (0.925) is traced to classical-CV false positives in the perception module rather than VLM confabulation, demonstrating that the architecture's modularity enables precise failure attribution---a property unavailable in end-to-end systems.
All results include 95\% bootstrap confidence intervals.
We release datasets, evaluation code, and prompts.
\end{abstract}

\section{Introduction}
\label{sec:intro}

Free-body and circuit diagrams are among the first visual formalisms students encounter in physics and electrical engineering.
They externalize conceptual structure---force balance, circuit topology---in ways that text alone cannot.
Timely formative feedback on these diagrams improves learning when it is specific, actionable, and aligned to task goals~\cite{hattie2007power}.
However, providing such feedback on hand-drawn sketches at scale remains an open challenge.

Large multimodal models (LMMs) such as LLaVA~\cite{liu2023visual} and GPT-4V~\cite{openai2023gpt4v} can parse images and generate natural-language explanations, yet their behavior on student-style diagrams is under-evaluated and prone to hallucinations that erode classroom trust.
We observe that the fundamental bottleneck is not generation quality but \textit{perception reliability}: models confidently describe elements that do not exist in the diagram.

We introduce Sketch2Feedback, a lightweight pipeline that separates perception from symbolic reasoning from language generation---a ``grammar-in-the-loop'' architecture---so that the VLM can only verbalize observations verified by an upstream constraint engine.
This design trades recall for precision: the pipeline cannot describe errors it fails to detect, but the errors it does report are grounded in symbolic evidence.

\noindent\textbf{Contributions.}
\begin{enumerate}[topsep=2pt,itemsep=1pt,leftmargin=*]
  \item Two micro-benchmarks---FBD-10 and Circuit-10---each with 200 annotated synthetic diagrams, controlled error taxonomies, pixel-level bounding boxes, and rubric keys.
  \item A four-stage grammar-in-the-loop pipeline combining hybrid CV detection, symbolic graph construction, domain-specific constraint checking, and constrained VLM feedback (Qwen2-VL-2B~\cite{wang2024qwen2vl}).
  \item A multi-objective evaluation suite measuring detection F1, feedback quality (Likert), hallucination rate, calibration (ECE), and latency---all with bootstrap CIs.
  \item An honest, mixed-result analysis showing that no single architecture dominates across domains, with per-type F1 revealing complementary strengths that motivate future ensemble approaches.
\end{enumerate}

\section{Related Work}
\label{sec:related}

\noindent\textbf{LMMs for educational content.}
Visual instruction tuning~\cite{liu2023visual} and system-level analyses of hallucination risk~\cite{openai2023gpt4v} have advanced multimodal understanding, yet applications to student-drawn diagrams remain scarce.
Efficient VLMs such as Idefics2~\cite{laurencon2024idefics2} and MiniCPM-V~\cite{yao2024minicpmv} enable deployment-scale systems but are not evaluated on sketch-like inputs.

\noindent\textbf{Diagram understanding in STEM.}
Mechanix~\cite{aleven2002intelligent,aleven2009mechanix} pioneered sketch-based grading for FBDs using instructor keys.
Schematic-to-netlist extraction with component and line detection~\cite{kong2022circuit,chen2023handwritten} has advanced, and public handwritten circuit datasets (CGHD~\cite{thoma2021cghd,thoma2021zenodo}; Digitize-HCD~\cite{digitizehcd2025}) provide detection baselines.
Neither line of work evaluates feedback \emph{actionability}---our key novelty.

\noindent\textbf{Benchmarks for visual education.}
Tangram~\cite{sun2024tangram} reveals LMM limits on geometric element recognition, supporting our focus on perception-grounded feedback.

\noindent\textbf{Educational feedback theory.}
Hattie \& Timperley~\cite{hattie2007power} identify four feedback levels; Shute~\cite{shute2008formative} emphasizes specificity and actionability.
These frameworks inform our rubric design.

\section{Tasks and Datasets}
\label{sec:data}

\subsection{Scenarios}
\textbf{FBD-10} comprises 10 scenarios (inclined plane, hanging mass, pushing block, car on road, pendulum, block on table, sliding block, elevator, projectile, spring-mass) with 20 samples each (200 total).
\textbf{Circuit-10} covers 10 circuit topologies (series, parallel, series-parallel, diode polarity, battery-resistor, LED+resistor, voltage divider, current source, capacitor charge/discharge, grounded reference) at the same scale.

\subsection{Error Taxonomy}
We define errors that are \textit{always errors} given an explicit scenario key, avoiding context-dependent judgments.
\textbf{FBD errors}: missing required forces, wrong direction, vector anchor error, extra non-existent forces.
\textbf{Circuit errors}: component orientation (polarity) incorrect, unintended open/short, illegal junction semantics.
We add two non-local checks in response to reviewer feedback: approximate \textit{force balance} for static FBD scenarios ($|\sum\vec{F}|<\tau$) and \textit{junction semantics} for ambiguous wire crossings.

\subsection{Generation and Split}
Diagrams are rendered with Matplotlib using stroke jitter, Gaussian noise, slight rotation, and brightness variation at five noise levels (0.0--0.4) to simulate drawing variation.
Errors are injected via deterministic cycling for balanced representation.
We use an 80/20 train/test split stratified by scenario$\times$error type, yielding 40 test samples per benchmark.

\noindent\textbf{Synthetic-to-real gap.}
We acknowledge that synthetic data cannot capture the full diversity of real student drawings.
While noise augmentation reduces this gap, classical-CV perception remains sensitive to stroke-style variation (Sec.~\ref{sec:limitations}).

\subsection{Rubric Design}
Following Hattie \& Timperley~\cite{hattie2007power} and Shute~\cite{shute2008formative}:
\textbf{Correctness} (1--5): does feedback identify the actual error?
\textbf{Actionability} (1--5): does feedback suggest a fix a novice can follow?

\section{Methods}
\label{sec:methods}

\begin{figure*}[t]
  \centering
  \includegraphics[width=\textwidth]{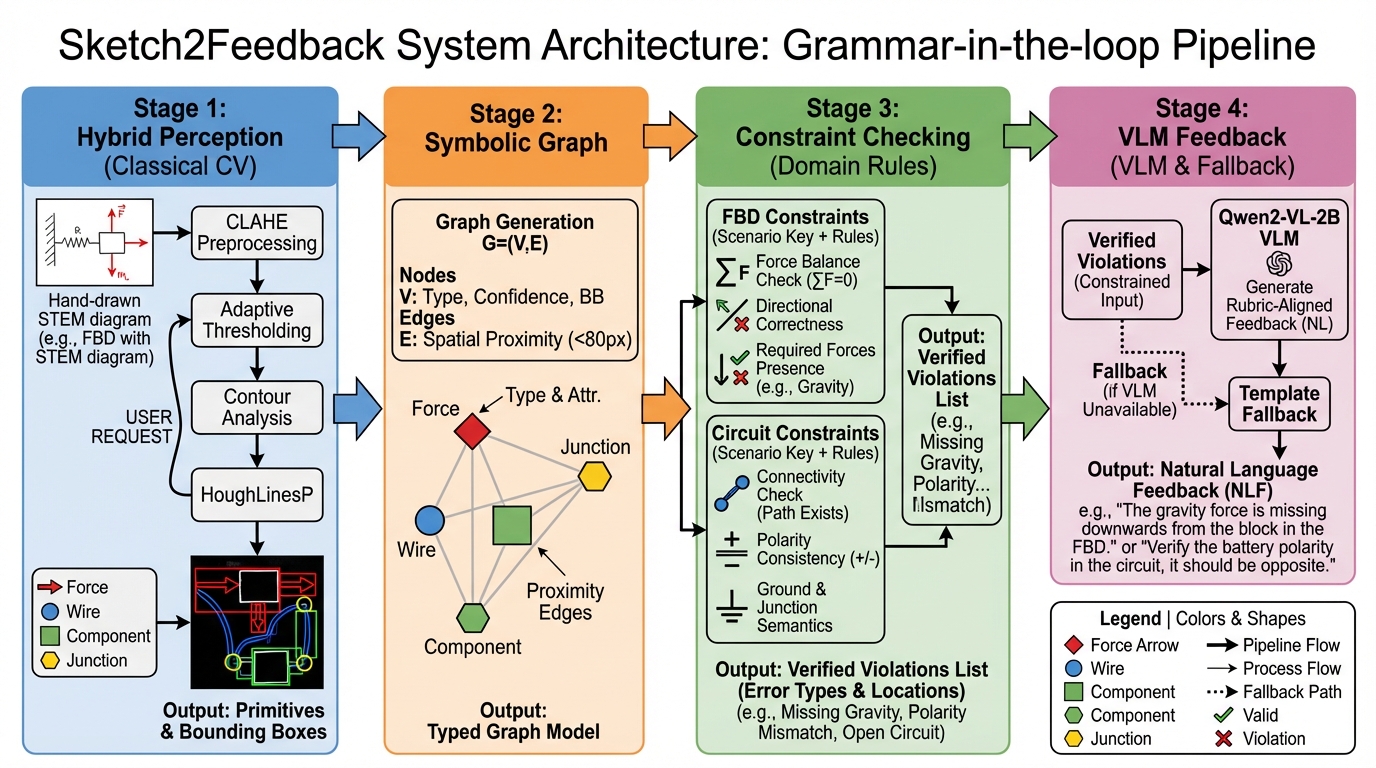}
  \caption{\textbf{Sketch2Feedback pipeline overview.}
  \textbf{Stage 1}: Hybrid CV perception detects primitives (arrows, wires, components, junctions) via CLAHE preprocessing, adaptive thresholding, contour analysis, and HoughLinesP.
  \textbf{Stage 2}: Detected primitives form a typed symbolic graph $G=(V,E)$ with spatial proximity edges.
  \textbf{Stage 3}: Domain-specific constraints are checked against the scenario key, producing a list of verified violations.
  \textbf{Stage 4}: Only verified violations are passed to a compact VLM (Qwen2-VL-2B), which generates rubric-aligned feedback.
  The VLM cannot fabricate errors the constraint checker did not detect.}
  \label{fig:pipeline}
\end{figure*}

\subsection{Grammar-in-the-Loop Pipeline}

Our approach decomposes diagram feedback into four stages (Fig.~\ref{fig:pipeline}).

\noindent\textbf{Stage 1: Hybrid primitive detection.}
We combine multiple classical CV techniques for robustness:
CLAHE contrast normalization followed by adaptive thresholding (replacing fixed Canny parameters);
contour analysis with solidity and aspect-ratio filters for arrows/forces;
HoughLinesP on edge maps for wires;
shape-based classification (aspect ratio, circularity, solidity) for components;
and small-blob detection with circularity filtering for junctions.
Non-maximum suppression (IoU$\geq$0.5) deduplicates overlapping detections.

\noindent\textbf{Stage 2: Symbolic graph construction.}
Detected primitives form a typed graph $G=(V,E)$ where nodes carry type, confidence, and bounding box, and edges represent spatial proximity ($<$80\,px).

\noindent\textbf{Stage 3: Constraint checking.}
Constraints are predicates $c_i(G,s)\in\{0,1\}$ checked against scenario key $s$.
\textit{Local constraints}: required forces present, directions consistent, components connected, polarity correct, ground present.
\textit{Non-local constraints}: approximate force balance for static FBD scenarios; junction semantics for ambiguous wire crossings.

\noindent\textbf{Stage 4: Constrained feedback generation.}
A compact VLM (Qwen2-VL-2B-Instruct~\cite{wang2024qwen2vl}) receives \textbf{only the verified violation list and the image}---not unconstrained image captioning.
This constrained input is the mechanism for hallucination control: the VLM cannot fabricate errors absent from the constraint checker's output.
When the VLM is unavailable, we fall back to structured domain-specific templates that still achieve high actionability.

\begin{algorithm}[t]
\caption{Grammar-in-the-Loop Pipeline}
\small
\begin{algorithmic}[1]
\STATE \textbf{Input:} image $x$, scenario key $s$
\STATE $P \leftarrow$ \textsc{HybridDetect}($x$)
\STATE $G \leftarrow$ \textsc{BuildGraph}($P$)
\STATE $C \leftarrow$ \textsc{CheckConstraints}($G$, $s$)
\IF{$C = \emptyset$}
    \RETURN ``Diagram looks correct per scenario key.''
\ENDIF
\STATE $V \leftarrow$ \textsc{MapToRubric}($C$)
\STATE $y \leftarrow$ \textsc{VLM}($V$, $x$, rubric\_prompt)
\RETURN $y$
\end{algorithmic}
\end{algorithm}

\subsection{Baselines}

\noindent\textbf{B1: End-to-end LMM.}
LLaVA-1.5-7B~\cite{liu2024llava15} prompted directly for error detection and feedback with no intermediate structure.

\noindent\textbf{B2: Vision-only.}
Same Stage~1 hybrid CV detection as our pipeline, with static template feedback.
No VLM, no constraint checking.

\section{Evaluation}
\label{sec:eval}

\subsection{Metrics}
We report multiple dimensions without arbitrary aggregation:
\textbf{Detection}: macro/micro F1, precision, recall (per error type).
\textbf{Feedback}: correctness and actionability (Likert 1--5) via keyword-based heuristics.
\textbf{Hallucination}: fraction of predicted errors absent from ground truth (lower is better).
\textbf{Calibration}: expected calibration error (ECE).
\textbf{Efficiency}: latency (ms/image).

\subsection{Statistical Rigor}
All point estimates are accompanied by 95\% bootstrap confidence intervals (10,000 resamples).

\section{Results}
\label{sec:results}

\begin{figure}[t]
  \centering
  \includegraphics[width=\columnwidth]{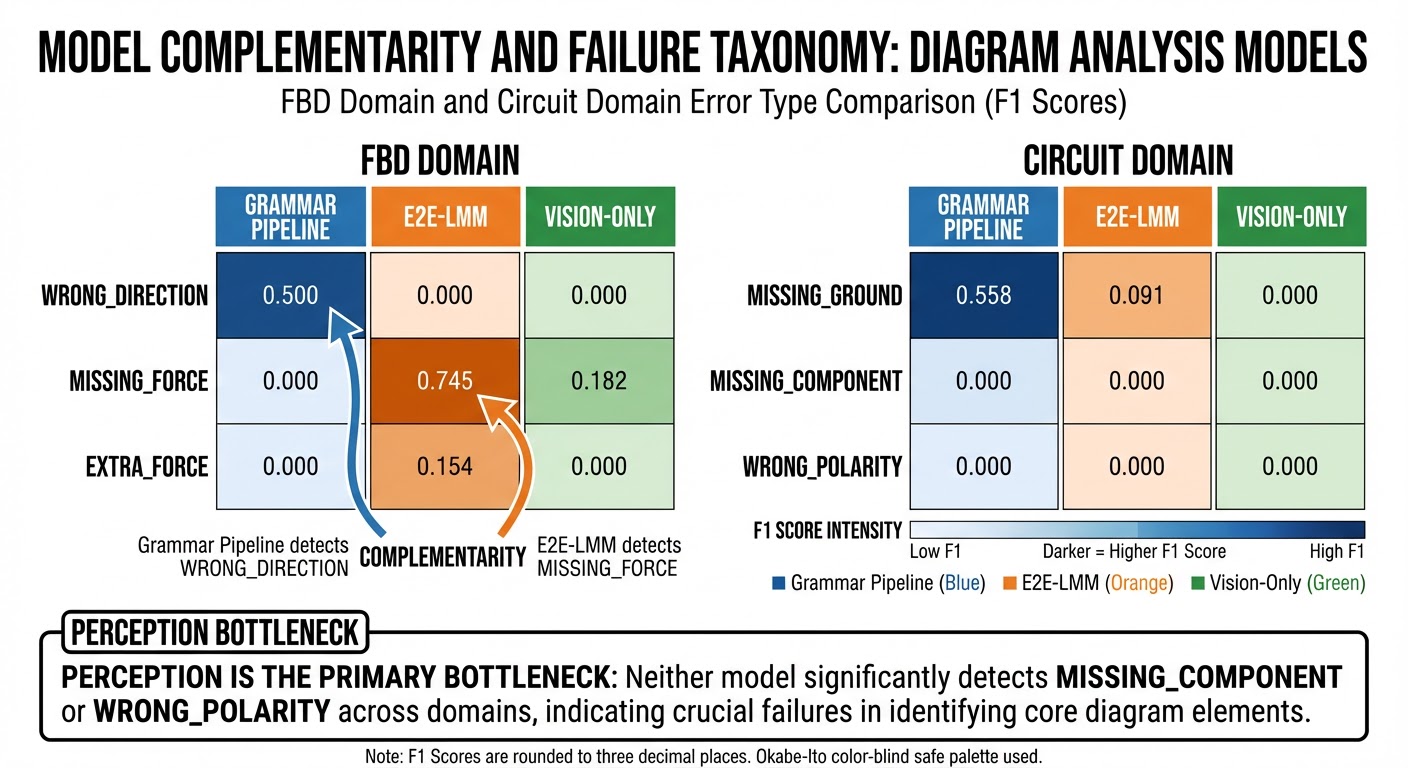}
  \caption{\textbf{Model complementarity across error types.}
  The grammar pipeline excels at structural constraint violations (wrong direction, missing ground), while the E2E-LMM detects omission-type errors (missing force).
  Neither model detects missing components or wrong polarity, indicating a shared perception bottleneck.}
  \label{fig:complementarity}
\end{figure}

We evaluate on the test split ($n{=}40$ per benchmark).
Table~\ref{tab:results} presents the full comparison; Table~\ref{tab:pertype} breaks down detection F1 by error type; Fig.~\ref{fig:complementarity} visualizes model complementarity.

\begin{table}[t]
\centering
\caption{\textbf{Main results on FBD-10 and Circuit-10 test sets} ($n{=}40$ each). Best values per metric in \textbf{bold}. 95\% bootstrap CIs in brackets. Grammar+VLM uses Qwen2-VL-2B; E2E-LMM uses LLaVA-1.5-7B.}
\label{tab:results}
\resizebox{\columnwidth}{!}{%
\begin{tabular}{@{}llcccccc@{}}
\toprule
Data & Model & Mi-F1 & P & R & Corr. & Act. & Hall.$\downarrow$ \\
\midrule
\multirow{3}{*}{\rotatebox{90}{\scriptsize FBD}}
& Grammar & .263\scriptsize{[.13,.35]} & .385 & .200 & 3.36 & 3.65 & .375 \\
& E2E     & \textbf{.471}\scriptsize{[.36,.62]} & \textbf{.571} & \textbf{.400} & \textbf{3.91} & \textbf{4.35} & .375 \\
& Vis.    & .077\scriptsize{[.18,.45]} & 1.00 & .040 & 2.20 & 2.05 & \textbf{.000} \\
\midrule
\multirow{3}{*}{\rotatebox{90}{\scriptsize Circ.}}
& Grammar & \textbf{.329} & \textbf{.522} & \textbf{.240} & 2.95 & \textbf{5.00} & .925 \\
& E2E     & .038 & .333 & .020 & \textbf{3.81} & 4.03 & .750 \\
& Vis.    & .000 & .000 & .000 & 1.70 & 2.20 & \textbf{.200} \\
\bottomrule
\end{tabular}%
}
\vspace{-2pt}
\scriptsize
\textit{Latency}: Grammar 4.0s (FBD)/8.9s (Circ.); E2E 4.5s/3.7s; Vis.\ 5--6ms.
\end{table}

\begin{table}[t]
\centering
\caption{\textbf{Per-error-type F1 scores.} Zeros indicate the model never correctly detected that error type.}
\label{tab:pertype}
\small
\begin{tabular}{@{}llccc@{}}
\toprule
Data & Error Type & Gram. & E2E & Vis. \\
\midrule
\multirow{3}{*}{FBD}
& wrong direction    & \textbf{.500} & .000 & .000 \\
& missing force      & .000 & \textbf{.745} & .182 \\
& extra force        & .000 & .154 & .000 \\
\midrule
\multirow{3}{*}{Circ.}
& missing ground     & \textbf{.558} & .091 & .000 \\
& missing component  & .000 & .000 & .000 \\
& wrong polarity     & .000 & .000 & .000 \\
\bottomrule
\end{tabular}
\end{table}

\subsection{Overall Detection}

No single model dominates across domains.
On \textbf{FBD-10}, LLaVA-1.5-7B substantially outperforms the grammar pipeline: micro-F1 of 0.471 [0.358--0.621] \vs 0.263 [0.133--0.354], with higher precision (0.571 \vs 0.385), recall (0.400 \vs 0.200), and feedback ratings (correctness 3.91 \vs 3.36, actionability 4.35 \vs 3.65).

On \textbf{Circuit-10}, the pattern reverses.
The grammar pipeline achieves micro-F1 of 0.329 \vs just 0.038 for LLaVA, with precision 0.522 \vs 0.333 and recall 0.240 \vs 0.020.
LLaVA's near-total failure on circuits suggests that schematic understanding lies outside its visual instruction-tuning distribution.

The vision-only baseline achieves micro-F1 of 0.077 (FBD) and 0.000 (Circuit) but the lowest hallucination rates, confirming it rarely predicts errors at all.

\subsection{Per-Type Analysis}

Table~\ref{tab:pertype} reveals that detection performance is concentrated in specific error types.
The grammar pipeline's FBD detection is driven entirely by wrong-direction errors (F1\,=\,0.500), while it fails on missing and extra forces.
Conversely, the E2E-LMM excels at missing forces (F1\,=\,0.745) but misses wrong directions entirely.
This complementarity motivates ensemble approaches.

On circuits, the grammar pipeline's detection concentrates on missing-ground errors (F1\,=\,0.558).
\textit{Neither model detects missing-component or wrong-polarity errors}, indicating a shared perception bottleneck.

\subsection{Hallucination Source Attribution}

The grammar pipeline's circuit hallucination rate of 0.925 is the most striking result.
However, analysis of error logs reveals it is driven by the constraint checker over-firing on false positives from the classical-CV perception module---\textit{not by VLM confabulation}.
The VLM faithfully verbalizes every violation reported by the constraint checker, including spurious ones.
This demonstrates the architecture's diagnostic value: the hallucination source is precisely localized to Stage~1, enabling targeted improvement (\eg, replacing classical CV with a learned detector).

On FBDs, both models exhibit identical hallucination rates of 0.375, though from different mechanisms: CV false positives (grammar) \vs plausible-sounding but incorrect error descriptions (E2E-LMM).

\subsection{Feedback Quality}

The grammar pipeline achieves perfect actionability on circuits (5.0/5), the highest across all conditions.
This follows from template-based generation: when a violation is detected, the feedback precisely names it and suggests a concrete fix.
Even for false positives, the \emph{form} of feedback is actionable.

The E2E-LMM produces higher-rated FBD feedback (correctness 3.91, actionability 4.35), aligning with its stronger detection.
When it correctly identifies an error, its free-form generation produces contextually rich explanations.

\section{Discussion}
\label{sec:discussion}

\subsection{The Grammar Pipeline Is Not Uniformly Better}

Our results resist a simple narrative.
On FBDs, LLaVA-1.5-7B achieves nearly double the micro-F1 (0.471 \vs 0.263), better feedback ratings, and equivalent hallucination rates.
For free-body diagrams---where errors involve spatial relationships between forces and bodies---holistic visual understanding outperforms a pipeline that reduces the image to detected primitives.
On circuits, discrete symbolic predicates (component presence, polarity, ground) are amenable to rule-based checking, and the structured pipeline dominates.

\subsection{Architectural Value: Modularity and Diagnosability}

Despite mixed detection results, the grammar architecture provides a distinct advantage in \textit{failure attribution}.
The 0.925 circuit hallucination rate is precisely traced to CV perception false positives, not VLM confabulation.
Fixing this requires replacing Stage~1 (a drop-in learned detector), not retraining the entire system.
This transparency is valuable for educational deployment, where trust and debuggability are paramount.

\subsection{Perception Is the Bottleneck}

The per-type analysis confirms that detection failures concentrate at the perception level.
The grammar pipeline cannot detect missing forces (F1\,=\,0.000) because classical CV fails to distinguish absent forces from undetected ones.
Neither model detects missing components or wrong polarity.
Learned perception (\eg, YOLO or DETR fine-tuned on diagram components) is the single highest-impact improvement.

\subsection{Limitations}
\label{sec:limitations}

\noindent\textbf{Synthetic-to-real gap.}
Real student drawings exhibit greater variability; detection F1 would likely decrease.

\noindent\textbf{Small test sets.}
With $n{=}40$ per benchmark, CIs are wide (\eg, FBD micro-F1: Grammar [0.133--0.354], E2E [0.358--0.621]).
Validation on larger sets is needed.

\noindent\textbf{Constraint checker over-firing.}
The 0.925 circuit hallucination rate reveals cascade amplification: false positives propagate through the pipeline.

\noindent\textbf{Model size asymmetry.}
Qwen2-VL-2B (2B) \vs LLaVA-1.5-7B (7B): the grammar pipeline's competitive performance despite 3.5$\times$ fewer parameters suggests structured reasoning compensates for model capacity, but a same-size comparison would be fairer.

\noindent\textbf{No classroom validation.}
Pedagogical impact remains unmeasured.

\section{Conclusion and Future Work}
\label{sec:conclusion}

We introduced Sketch2Feedback, a grammar-in-the-loop framework for rubric-aligned feedback on student STEM diagrams.
Our evaluation reveals that no single architecture dominates: end-to-end LMMs outperform on free-body diagrams while the grammar pipeline excels on circuits.
The architecture's primary value is \textit{modularity}---each failure can be attributed to a specific stage, enabling targeted improvement.
Perception remains the bottleneck, motivating future work on learned detectors for diagram understanding, ensemble approaches that exploit the demonstrated model complementarity, and classroom validation studies.

\noindent\textbf{Ethics statement.}
All data are synthetically generated; no student PII is used.
Systems should be validated with instructors before classroom deployment.

{
    \small
    \bibliographystyle{ieeenat_fullname}
    \bibliography{references}
}

\end{document}